\documentclass[letterpaper, 12pt]{ewshm}

\usepackage[dvips,final]{graphicx}

\usepackage{makecell}
\usepackage{enumerate}
\usepackage{color}
\usepackage{subfig}
\usepackage{caption}
\usepackage{amsfonts}
\usepackage{amsmath}
\usepackage{amssymb}
\usepackage{times}
\usepackage{cite}
\usepackage{hhline}
\usepackage{compactbib}
\usepackage{graphicx}        
\usepackage{amsmath,amsfonts,amssymb}
\usepackage{geometry}
\usepackage{lineno,hyperref}
\usepackage{mathrsfs}
\usepackage[labelsep=period]{caption}
\usepackage{tikz}
\usetikzlibrary{shapes,arrows}
\usetikzlibrary{bayesnet}
\usetikzlibrary{fit,positioning}
\usepackage{pdfpages}

\definecolor{halfgreen}{RGB}{0,128,0}
\definecolor{ahsred}{RGB}{192,0,0}

\renewcommand{\thefigure}{\arabic{figure}}
\renewcommand{\thetable}{\Roman{table}}

\newcommand{\beq}{\begin{equation}}
\newcommand{\eeq}{\end{equation}}
\newcommand{\bgqar}{\begin{eqnarray}}
\newcommand{\enqar}{\end{eqnarray}}
\newcommand{\bgqarn}{\begin{eqnarray*}}
\newcommand{\enqarn}{\end{eqnarray*}}
\newcommand{\bgary}{\begin{array}}
\newcommand{\enary}{\end{array}}

\long\def\symbolfootnote[#1]#2{\begingroup%
\def\thefootnote{\fnsymbol{footnote}}\footnote[#1]{#2}\endgroup}

\makeatletter
\renewcommand\@biblabel[1]{#1.}
\makeatother

\begin{document}

%----------------------------------------------------------------------------------------------------------------------
% COVER PAGE
%----------------------------------------------------------------------------------------------------------------------

\vspace*{4.4cm}

\noindent Title: \textbf{A decision framework for selecting information-transfer strategies in population-based SHM}

\vspace{1.6cm}

\noindent
$
\begin{array}{ll}
\text{Authors}: 
& \text{Aidan J.\ Hughes$^1$}  \\ 
& \text{Jack Poole$^1$} \\
& \text{Nikolaos Dervilis$^1$} \\
& \text{Paul Gardner$^{1,2}$} \\
& \text{Keith Worden$^1$} \\

\end{array}
$

\newpage

%----------------------------------------------------------------------------------------------------------------------
% 1st PAGE
%----------------------------------------------------------------------------------------------------------------------

\vspace*{60mm}

%----------------------------------------------------------------------------------------------------------------------
\noindent \uppercase{\textbf{Abstract}} \vspace{12pt} 

Decision-support for the operation and maintenance of structures provides significant motivation for the development and implementation of structural health monitoring (SHM) systems. Unfortunately, the limited availability of labelled training data hinders the development of the statistical models on which these decision-support systems rely. Population-based SHM seeks to mitigate the impact of data scarcity by using transfer learning techniques to share information between individual structures within a population. The current paper proposes a decision framework for selecting transfer strategies based upon a novel concept -- the expected value of information transfer -- such that negative transfer is avoided. By avoiding negative transfer, and by optimising information transfer strategies using the transfer-decision framework, one can reduce the costs associated with operating and maintaining structures, and improve safety.

%----------------------------------------------------------------------------------------------------------------------

\symbolfootnote[0]{\hspace*{-7mm} Aidan J.\ Hughes, Email: aidan.j.hughes@sheffield.ac.uk. $^1$Dynamics Research Group, Department of Mechanical Engineering, University of Sheffield, Sheffield, S1 3JD, UK. $^2$Frazer-Nash Consultancy, Warrington, UK.}

%------------------------------------------------------------------------------------------------------------
% SECTION 1: INTRODUCTION
%------------------------------------------------------------------------------------------------------------

\vspace{12pt} 
\noindent \uppercase{\textbf{Introduction}}  \vspace{12pt} 

%\noindent \uppercase{\textbf{Battery}}  \vspace{12pt} 

Structural health monitoring (SHM) systems provide a means of augmenting operation and maintenance decision processes with up-to-date information regarding the health-state of a structure or system \cite{Farrar2013}. In order to assign features extracted from sensor data to meaningful categories in the context of the decision process (e.g. damage extent and locations), SHM systems rely on statistical classification models. Such models are typically learned from data; however, this can be challenging in many SHM applications, as data labelled with contextual information can be prohibitively expensive, or otherwise infeasible to obtain.

Population-based structural health monitoring (PBSHM), seeks to address some of the limitations associated with data scarcity that arise in traditional SHM \cite{Bull2021,Gosliga2021,Gardner2021b,Tsialiamanis2021}. A tenet of the population-based approach to SHM is that information can be shared between sufficiently-similar structures in order to improve predictive models. Transfer learning techniques, such as domain adaptation, have been shown to be a highly-useful technology for sharing information between structures when developing statistical classifiers for PBSHM \cite{Gardner2022kernel,Poole2022alignment,Gardner2022aircraft}. Nonetheless, transfer-learning techniques are not without their pitfalls. In some circumstances, for example if the data distributions associated with the structures within a population are dissimilar, applying transfer learning methods can actually be detrimental to classification performance – this phenomenon is known as \textit{negative transfer} \cite{Wang2019negative}. When considered in the context of operation and maintenance decision processes, negative transfer has significant implications. Deterioration in classification performance could translate to unnecessary inspections or repairs, and even critical maintenance interventions being missed entirely. Such changes in operation and maintenance strategy would result in additional costs being incurred and could undermine the integrity and safety of structures. Given the potentially-severe consequences of negative transfer, it is prudent for engineers to ask questions such as “when should one transfer information between structures?” and ``from where should information be transferred?" \cite{Pan2010survey}.

In an attempt to offer guidance on how one can answer the aforementioned questions in a principled manner, the current paper aims to highlight the models and information a decision-making agent requires in order to determine whether transfer can be beneficial. Moreover, a decision-theoretic framework for selecting information-transfer strategies is presented. The proposed framework centres around the notion of \textit{expected value of information} (EVoI). In general, the EVoI captures the difference in the maximum expected utility that can be achieved for a decision process when considering both the absence and presence of said information. In the context of transfer learning, the notion of EVoI becomes useful as it can be used to predict the change in utility achieved for structural operation and maintenance decisions when using a model subject to differing transfer strategies. Here, it is worth recognising that assessing the expected change in expected utility between doing no transfer and a given transfer strategy is equivalent to assessing the risk of negative transfer.

The remainder of the current paper is structured as follows. Section 2 provides some background theory on the use of transfer learning techniques in PBSHM. Section 3 provides background theory on decision theory and value of information. Section 4 presents the decision framework for selecting transfer strategies in PBSHM and highlights the key information, metrics and models required to make such decisions. Section 5 offers some discussion around how the parameters of the decision framework can be tailored according to the nature of the PBSHM problem one is trying to solve. Finally, a summary is given in Section 6.

%------------------------------------------------------------------------------------------------------------
% SECTION 2:
%------------------------------------------------------------------------------------------------------------

\vspace{20pt}

\noindent \uppercase{\textbf{Transfer Learning for PBSHM}} \vspace{12pt}

Transfer-learning techniques seek to use information from a data-, or label-, rich \textit{source domain} $\mathcal{D}_s = \{ \mathbf{x}_{s,i}, y_{s,i} \}_{i=1}^{N}$, in order to improve predictions in a data-, or label-, scarce \textit{target domain} $\mathcal{D}_t = \{\mathbf{x}_{t,j},y_{t,j}\}_{j = 1}^{M}$\footnote{Here, it is worth noting that for the target domain it may be the case that only $\mathbf{x}_t$ are available, or that only a small subset of $\mathbf{x}_t$ are labelled.}. In PBSHM, it is assumed that there will be some subset of individual structures within a population for which data are available with contextual labels corresponding to damage states and operational conditions that are pertinent to O\&M decision processes. These structures with labelled data available can be treated as source domains. It then follows that structures within a population for which data, or labels, are unavailable can be considered as target domains \cite{Gardner2021b}. In general, it is assumed that there are some differences between source and targets domains; in particular, it is assumed that the marginal distributions of observable data differ, i.e.\ $P(X_s) \neq P(X_t)$, and/or that the conditional distributions of labels differ, i.e.\ $P(\mathbf{y}_s|X_s) \neq P(\mathbf{y}_t|X_t)$. In the context of a population of structures, these discrepancies between source and targets domains arise because of factors such as manufacturing variability; geometric, topological, and material differences between structures; and operational and environmental differences.

Several transfer-learning approaches have been applied to PBSHM. Various flavours of \textit{domain adaptation} have been used to harmonise source and target domains in the context of PBSHM. In \cite{Gardner2018adaptation}, transfer component analysis (TCA), joint domain adaptation (JDA), and adaptation regularisation-based transfer learning (ARTL) are applied to several PBSHM case studies. In \cite{Poole2022alignment}, statistic alignment with domain adaptation and partial domain adaptation is successfully demonstrated on various engineering case studies, including transfer between the Z24 and KW51 bridges. Balanced distribution adaptation (BDA) is demonstrated in \cite{Gardner2022aircraft}. Kernelised Bayesian transfer learning (KBTL) is applied in \cite{Gardner2022kernel}. Neural approaches to domain adaptation are applied to a population of rotating machines and a population of lattice structures in \cite{Yu2020adaptation} and \cite{Soleimani2023zeroshot}, respectively. In addition to domain adaptation, other transfer learning approaches have been applied in SHM including hierarchical Bayesian modelling \cite{Bull2023hierarchical} and fine-tuning \cite{Cao2018preprocessing}.

\vspace{12pt} 
\noindent \textbf{Negative Transfer}  \vspace{12pt} 

An important consideration when applying transfer learning techniques in PBSHM is the possibility of negative transfer \cite{Wang2019negative}. Negative transfer is characterised by a degradation in predictive performance in the target domain post-transfer. Negative transfer can occur if the source joint distribution $P(X_s,\mathbf{y}_s)$ and target joint distribution $P(X_t, \mathbf{y}_t)$ are not sufficiently similar; or, if the algorithm used to conduct transfer is unable to find the correct mapping for other reasons, such as non-uniqueness of solutions. 

Negative transfer has severe implications in the context of structural health monitoring and asset management. By definition, negative transfer results in erroneous classifications. In the most benign cases, these misclassifications may just be between two undamaged classes under different environmental conditions (e.g. normal and cold temperature). However, if the misclassifications occur between an undamaged class and a damage class, then at best an unnecessary action will be taken, and at worst a critical intervention will be missed leading to a catastrophic failure of a structure of interest. In either case, misclassifications incur some cost when framed in the context of an operation and maintenance decision process; therefore, it is discerning to anticipate and avoid negative transfer whenever possible.

%------------------------------------------------------------------------------------------------------------
% SECTION 2:
%------------------------------------------------------------------------------------------------------------

\vspace{24pt}
\noindent \uppercase{\textbf{Decision-making for SHM}} \vspace{12pt}

As alluded to in the introduction, one of the primary motivations for the development and implementation of structural health monitoring systems is decision support for operation and maintenance planning. Integration of SHM information O\&M decisions processes have been demonstrated in \cite{Nielsen2013,Schobi2016,Hughes2021,Kamariotis2022}. 

Decidable actions relating to the O\&M of structures broadly fall into two categories; interventional actions, and observational actions. The result of an interventional action, such as the replacement/repair of a component in a structure, corresponds to a modification of a future/current state of interest (e.g.\ structural health state), via some causal mechanism. On the other hand, outcomes of observational actions, such as performing an inspection of a structure, do not \textit{cause} an actual change in a state of interest, but rather influence a decison-making agent's belief regarding a state of interest. 

A decision-theoretic approach to selecting interventional actions involves two key aspects; reasoning under uncertainty, and maximisation of expected utility. As outlined in \cite{Kjaerulff2008}, both of these aspects are captured in a form of probabilistic graphical model known as \textit{influence diagrams}. Within influence diagrams, graph nodes are used to represent random variables, decidable actions and cost/utilities. Graph edges are then used to specify conditional dependencies between nodes. From the dependencies specified by the graphical element, influence diagrams are parametrised using marginal and conditional probability distributions, and cost/utility functions. Reasoning can be accomplished using influence diagrams by conditioning on observable variables and decidable actions, and using inference algorithms to compute updated beliefs for the remaining unobserved variables. Expected utilities can be computed by taking the product of the posterior beliefs over random variables with their corresponding utility functions. Subsequently, optimal decisions can be made by selecting actions such that expected utilities are maximised. 

Whereas interventional actions are selected based on maximum expected utility, observational actions are selected based upon \textit{expected value of information}. An intuitive definition of expected value of information is `the price an agent would be willing to pay in order to gain access to information regarding an otherwise unknown or uncertain state, prior to making a decision'. The expected value of information is computed as the difference in maximum expected utility achievable for an interventional decision process when considered with, and without, the presence of the said information. 

To elucidate this definition further, one can consider the scenario presented in \cite{Hughes2022,Hughes2022enhanced}, where the expected value of information from inspection is considered. In this example, an agent must decide whether to perform maintenance based on a probabilistic health-state prediction from a statistical classifier. To support this interventional decision, the agent also decides whether or not to inspect the structure in order to gain knowledge of the current health state at some additional cost. If there is sufficient uncertainty in the classifier prediction about whether the structure is healthy or damaged (and consequently, a high degree of uncertainty in what the optimal decision is), then the expected value of performing an inspection will exceed the cost of conducting the inspection and the optimal strategy is to obtain further information prior to performing an intervention, in order to avoid (potential) catastrophic failure. 

To summarise, decisions in the context of SHM can be categorised as either interventional, or observational. Interventional decisions as selected by maximising expected utility, whereas observational decisions are made based upon value of information.

%------------------------------------------------------------------------------------------------------------

%

%\vspace{12pt}
%\noindent \textbf{VFP model}  \vspace{12pt} 

%\end{subequations}
%

%------------------------------------------------------------------------------------------------------------
%
%------------------------------------------------------------------------------------------------------------
\newpage
\vspace{24pt}
\noindent \uppercase{\textbf{Transfer Learning Decision Framework}} \vspace{12pt} 

As with structural inspections, information transfer between two structures within a population does not cause a change in a structure's states of interest; as such, it is natural to consider decisions regarding transfer as being of the observational type. In order to decide on a transfer strategy, the concept of \textit{expected value of information transfer} (EVIT) is introduced.

In order to define EVIT, and to form a transfer-strategy decision process around it, here, it is useful to establish some notation. The context for a transfer decision process $\mathcal{I}_{\mathcal{T}}$ can be defined as follows:

\begin{itemize}
	\item a set of candidate transfer strategies $\mathscr{T} = \{ \mathcal{T}_{\tau} \}_{\tau=0}^{N_{\mathcal{T}}}$;
	\item a target domain $\mathcal{D}_t$;
	\item a set of candidate source domains $\mathscr{D}_s = \{ \mathcal{D}_{s,k}\}_{k=1}^{N_s}$;
	\item a prediction task in the target domain $P(\mathbf{y}_t|X_t, \mathcal{T})$;
	\item a decision process on the target domain $\mathcal{I}_{t}$ that is dependent on $\mathbf{y}_t$;
	\item a set of prediction-quality measures (e.g.\ prediction accuracy) $\mathcal{Q}(\hat{\mathbf{y}}_t,\mathbf{y}_t^{\ast})$, where $\hat{\mathbf{y}}_t$ and $\mathbf{y}_t^{\ast}$ are the predicted and true targets, respectively;
	% \item a set of similarity measures $\mathscr{S} = \{ \mathcal{S}_{\sigma} \}_{\sigma=1}^{N_\mathcal{S}}$;
	\item a utility function $U(\mathcal{Q})$, specified with respect to $\mathcal{I}_{t}$;
	\item a utility function $U(\mathscr{T})$ that specifies the costs of performing different transfer strategies.
\end{itemize}

The transfer strategies that are the subject of a transfer decision process $\mathcal{I}_{\mathcal{T}}$, for a given target domain and prediction task, can be defined as a tuple $\mathcal{T}(\mathscr{D}_{s,\tau}, \mathcal{A}) \in \mathscr{T}$, where $\mathcal{A}$ is a transfer-learning algorithm from a set of candidate algorithms $\mathscr{A} = \{ \mathcal{A}_a \}_{a=0}^{N_{\mathcal{A}}}$, and $\mathscr{D}_{s,\tau} \subseteq \mathscr{D}_s$. Here, it is also useful to define a null transfer strategy, corresponding to the case for which no transfer is performed, $\mathcal{T}_0 = \mathcal{T}(\mathscr{D}_{s,0} = \emptyset,\mathcal{A}_0 = \mathbb{I})$, where $\mathbb{I}$ represents an identity operation.

With some of the machinery now established, one can now define EVIT as the price a decision-maker should be willing to pay to transfer information from source domains to a target domain, prior to making a decision using predictions in the target domain. More specifically, the EVIT of a specific transfer strategy $\mathcal{T}$ can be expressed as,

\begin{equation}
	\text{EVIT}(\mathcal{T}) = EU(\mathcal{Q}|\mathcal{T}) - EU(\mathcal{Q}|\mathcal{T}_0) = P(\mathcal{Q}|\mathcal{T})\cdot U(\mathcal{Q}) - P(\mathcal{Q}|\mathcal{T}_0)\cdot U(\mathcal{Q}),
	\label{eq:EVIT}
\end{equation}
\noindent where $EU$ denotes the expected utility.

Equation (\ref{eq:EVIT}) implies that $\text{EVIT}(\mathcal{T}_0) = 0$. Additionally, $\text{EVIT}(\mathcal{T}) < 0$ and $\text{EVIT}(\mathcal{T}) > 0$ correspond to negative and positive transfer, respectively, in the context of $\mathcal{I}_t$. Furthermore, an optimal transfer strategy can now be defined,

\begin{equation}
	\mathcal{T}^{\ast} = \mathcal{T}(\mathscr{D}_{s,\tau}^{\ast}, \mathcal{A}^{\ast}) = \underset{\mathcal{T}}{\text{argmax}} \big[ \text{EVIT}(\mathcal{T}) + U(\mathcal{T}) \big].
	\label{eq:T_opt}
\end{equation}

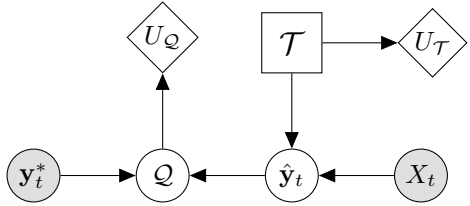
\begin{figure}[t!]
	\centering
	\begin{tikzpicture}[x=1.7cm,y=1.8cm]
		
		% Nodes for plate GM
		\node[det] (UQ) {$U_{\mathcal{Q}}$} ;
		\node[latent,below=1cm of UQ] (Q) {$\mathcal{Q}$} ;
		\node[obs, left=1cm of Q] (y_true) {$\mathbf{y}_t^{\ast}$} ;
		\node[latent, right=1cm of Q] (y_pred) {$\hat{\mathbf{y}}_{t}$} ;
		\node[obs, right=1cm of y_pred] (X_t) {$X_t$} ;
		\node[rectangle,draw=black,minimum width=0.8cm,minimum height=0.8cm,above=1cm of y_pred] (T) {$\mathcal{T}$} ;
		\node[det, right=1cm of T] (UT) {$U_{\mathcal{T}}$} ;
		
		\edge {Q} {UQ} ; %
		\edge {y_true,y_pred} {Q} ; %
		\edge {X_t,T} {y_pred} ; %
		\edge {T} {UT} ; %

	\end{tikzpicture}
	\caption{An influence diagram representing a transfer decision process. Random variables are represented as circular nodes, with observed variables shaded. Decisions are represented with square nodes. Utility/cost functions are represented with rhombic nodes.}
	\label{fig:TransferID1}
\end{figure}

The optimisation task shown in equation (\ref{eq:T_opt}) captures the key step for selecting information-transfer strategies in a decision-theoretic manner. To further elucidate the computation of EVIT and the associated optimisation, it is helpful to define an influence diagram \cite{Kjaerulff2008} representing the transfer decision process $\mathcal{I}_\mathcal{T}$. A general example for $\mathcal{I}_{\mathcal{T}}$ is shown in Figure \ref{fig:TransferID1}. There are several key dependencies highlighted by Figure \ref{fig:TransferID1}. Firstly, it can be seen in the graphical model that $\mathcal{Q}$ and $\mathcal{T}$ have utility/cost functions associated with them, denoted by the rhombic nodes. It is also shown in Figure \ref{fig:TransferID1}, that the predictions in the target domain $\hat{\mathbf{y}}_t$ are conditionally dependent on the decidable transfer strategy $\mathcal{T}$. The prediction quality $\mathcal{Q}$ is conditionally dependent on both the predicted targets and the true targets $\mathbf{y}_t^{\ast}$. Finally, it is evident from Figure \ref{fig:TransferID1} that the prediction quality $\mathcal{Q}$ is conditionally dependent on the transfer strategy $\mathcal{T}$, albeit indirectly via the predicted targets $\hat{\mathbf{y}}_t$; thus demonstrating that the expected utility of prediction performance $EU(\mathcal{Q})$ is dependent on $\mathcal{T}$, as stated in Equation (\ref{eq:EVIT}).

At this stage, the keen-eyed reader will notice a seemingly-fatal issue with the transfer decision process outlined in Figure \ref{fig:TransferID1}. The issue, of course, being that (by definition) the true prediction targets $\mathbf{y}_t^{\ast}$ are unavailable for the target domain $\mathcal{D}_t$, thereby preventing the computation of $P(\mathcal{Q}|\mathcal{T})$. To circumvent this obstacle, one can exploit a foundational concept in PBSHM to reframe the prediction $P(\mathcal{Q}|\mathcal{T})$ -- namely, the structural similarity between members of a population. It is proposed in \cite{Bull2021,Gosliga2021,Gardner2021b} that transfer between structures that are similar in terms of aspects such as their topology, geometry, and materials should yield superior results compared to transfers between structures that are dissimilar.

In \cite{Poole2023negative}, a methodology is presented that allows one to obtain a probability distribution over a prediction-quality measure following transfer from a single source domain to a target domain, using a measure of similarity between the source domain structure and target domain structure as an input. In particular, this methodology is demonstrated to predict damage-classification accuracy for a target domain in a population of 10 degree-of-freedom mass-spring systems with differing boundary conditions. This was accomplished by training a beta-likelihood Gaussian process to regress from the modal assurance criterion (a proxy for structural similarity) to the classification accuracy. The training data for this model were generated by repeatedly selecting pairs of structures from the available source domains, obscuring the class labels for one domain to create a pseudo-target domain, then using the other source domain to perform a transfer task onto the pseudo-target domain. Cross-referencing the post-transfer predictions with the labels for the pseudo-target domain yields a classification accuracy which, in combination with the value of modal assurance criterion between the two structures, gives data on which a regression model can be learned.

Here, the current paper proposes adopting a similar approach for obtaining probability distributions over $\mathcal{Q}$ whereby a set of similarity measures are introduced, $\mathscr{S} = \{ \mathcal{S}_{\sigma} \}_{\sigma=1}^{N_{\mathcal{S}}}$ where $\mathcal{S}(\mathcal{R}_{a},\mathcal{R}_{\mathbf{b}})$ captures the similarity between a target structure $a$ and a set of source structures $\mathbf{b}$ given some common representation approach $\mathcal{R}$. In \cite{Poole2023negative}, modeshapes are used as a representation $\mathcal{R}$ for a structure, with the MAC used as the similarity score $\mathcal{S}$. In \cite{Gosliga2021}, structures are represented as graphs and the Jaccard index is used as a measure of similarity.

Here, it is suggested that $N_{\mathcal{A}}$ models $P(\mathcal{Q}_a|\mathcal{S},\mathcal{T})$ (or more explicitly, $P(\mathcal{Q}_a|\mathcal{S},\mathcal{A}_a ; \mathscr{D}_{s,\tau})$) are learned from training data $ \{ \{ \mathcal{S}_i,\mathcal{Q}_{a,i} \}_{i=1}^{N_{\text{train}}}\}_{a=1}^{N_{\mathcal{A}}}$, treating $\mathcal{S}_i$ as inputs, and $\mathcal{Q}_i$ as the corresponding targets. Training examples of $\mathcal{Q}_{a,i}$ can be obtained by repeatedly applying each candidate transfer-learning algorithm $\mathcal{A}_a$ to transfer from a subset of source domains $\mathscr{D}_{s,\text{sub}}$ to a pseudo-target domain $\mathcal{D}_p$, differing $\{\mathscr{D}_{s,\text{sub}},\mathcal{D}_p \}$ for each of the $N_{\text{train}}$ repetitions, where $\mathscr{D}_{s,\text{sub}},\mathcal{D}_p \in \mathscr{D}_s$, but $\mathcal{D}_p \notin \mathscr{D}_{s,\text{sub}}$. Corresponding $\mathcal{S}_i$ can be obtained by assessing $\mathcal{S}(\mathcal{R}_p,\mathcal{R}_{\text{sub}})$. 

\begin{figure}[t!]
	\centering
	\begin{tikzpicture}[x=1.7cm,y=1.8cm]
		
		% Nodes for plate GM
		\node[det] (UQ) {$U_{\mathcal{Q}}$} ;
		\node[latent,below=2.5cm of UQ] (Q) {$\mathcal{Q}$} ;
		\node[latent, right=1cm of Q] (S) {$\mathcal{S}$} ;
		\node[latent, above=1cm of S] (Rs) {$\mathcal{R}_s$} ;
		\node[obs, right=1cm of Rs] (R) {$\mathcal{R}_t$} ;
		\node[rectangle,draw=black,minimum width=0.8cm,minimum height=0.8cm,above=1cm of Rs] (T) {$\mathcal{T}$} ;
		\node[det, right=1cm of T] (UT) {$U_{\mathcal{T}}$} ;

		\edge {Q} {UQ} ; %
		\edge {T,S} {Q} ; %
		\edge {R,Rs} {S} ; %
		\edge {T} {UT,Rs} ; %

	\end{tikzpicture}
	\caption{An influence diagram representing a transfer decision process reframed in terms of structural similarity between source and target domains.}
	\label{fig:TransferID2}
\end{figure}
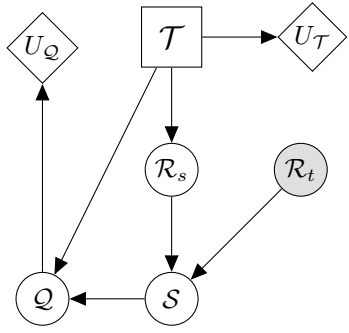

Provided that the model $P(\mathcal{Q}|\mathcal{S},\mathcal{T})$ can be learned from data generated using $\mathscr{D}_s$, the decision process shown in Figure \ref{fig:TransferID1} can be replaced with that shown in Figure \ref{fig:TransferID2}. In Figure \ref{fig:TransferID2}, the dependencies between $\mathcal{Q}$, $\mathcal{T}$, and their respective utility/cost functions are unchanged from Figure \ref{fig:TransferID1}. In Figure \ref{fig:TransferID2}, however, $Q$ is dependent on $\mathcal{S}$ and $\mathcal{T}$, as it is assumed that $Q$ is a function of both the similarity between source and target domains structures, and the transfer algorithm $\mathcal{A}$ associated with a given transfer strategy $T$. Here, it is also worth recognising from Figure \ref{fig:TransferID2}, that one no longer needs to know $X_t$ or $\mathbf{y}_t^{\ast}$. In their place, a representation of the target-domain structure is considered to be observed. 

To summarise, the reframed transfer decision process shown in Figure \ref{fig:TransferID2} does not fundamentally change the optimisation steps given in equations (\ref{eq:EVIT}) and (\ref{eq:T_opt}), as the ultimate goal is still to find the optimal algorithm $\mathcal{A}^{\ast}$ and optimal set of source domains $\mathscr{D}_{s,\tau}^{\ast}$. Rather, it reparametrises the conditonal distribution for $\mathcal{Q}$ in terms of a variable that is available/observable for given source and targets domains -- the structural similarities. Nonetheless, there are some limitations and considerations for the approach outlined in the current section -- these topics are discussed in the next section.

\vspace{20pt}
\newpage
\noindent \uppercase{\textbf{Discussion}} \vspace{12pt} 

The transfer-decision framework outlined in the previous section provides an approach for selecting transfer strategies such that predictions in the target domain are optimised for making decisions regarding operation and maintenance. There are, however, some limitations and considerations that should be highlighted and discussed.

Firstly, the approach requires the number of candidate source domains $N_s$ to be greater than 1. If $N_s \leq 1$, then training data cannot be generated, meaning $P(\mathcal{Q}|{S},\mathcal{T})$ cannot be learnt and the expected utility $EU(\mathcal{Q}|\mathcal{T})$ cannot be assessed. For such cases, an alternative decision framework would have to be used. A formulation for this alternative decision process is left as future work. 

Additionally, while the framework presented is intended to be quite general, in practice, generality comes at the price of inflated computational costs. One part of the general decision model that is computationally expensive is the generation of the training data. In this stage, depending on the size of the set of candidate source domains, a very large number of transfer tasks may be generated, bounded only by $N_s\cdot2^{N_s-1}$ per transfer algorithm $a$\footnote{For each of the $N_s$ possible $\mathcal{D}_p$, the number of possible source domain subsets corresponds to the cardinality of the power set of $\mathscr{D}_s$ with $\mathcal{D}_p$ removed, $2^{N_s-1}$.}. To curtail the computational expense, one could introduce constraints into the decision problem. For example, limiting $\mathscr{T}$ to only include single-source domain transfer strategies would cap the number of transfers in the data generation stage to $N_s(N_s-1)$. Alternatively, one could introduce heuristic stopping criteria into the data generation stage; for example, using a limit of 100 randomly-selected transfer strategies.  Further computational expense is incurred when considering numerous transfer learning algorithms, both in the learning and prediction stages. These expenses can be managed by limiting the number of candidate transfer algorithms $N_{\mathcal{A}}$.

Finally, a key challenge with implementing the transfer decision framework presented lies in the selection and modelling of prediction-quality measures $Q$. It is highly important that $Q$ are meaningful in the context of structural operation and maintenance decision processes, as one must be able to assign value/costs to these measures. One fairly simple example of measures that have direct interpretability with respect to structural operation and maintenance are type-I (false positive), and type-II (false negative) error rates. In the context of SHM, type-I errors correspond to damage being identified in cases where there is none, typically leading to costs associated with unnecessary inspections. Type-II errors, on the other hand, correspond to damage that is present being missed, which could potentially result in high costs from catastrophic structural failures. Oftentimes, these quality measures will be constrained and/or correlated in some way, meaning novel regression techniques may be required for learning $P(\mathcal{Q}|\mathcal{S},\mathcal{T})$. The investigations of appropriate prediction-quality measures and modelling techniques are also left as future work.

%Performance scores should be meaningful in the context of decision-making. simple - e.g type-I, type-II error rates. more complicated - pairwise misclassifications with each misclassication cost associated with the difference in MEU given class and optimal action

%The relationship between the natural frequencies and temperature is highly non-linear which is in contrast to lower natural frequencies \cite{hios2009stochastic}.  

%------------------------------------------------------------------------------------------------------------
%                                          CONCLUDING REMARKS
%------------------------------------------------------------------------------------------------------------
\vspace{24pt}

\noindent \uppercase{\textbf{Concluding Remarks}} \vspace{12pt}

To conclude, adopting a population-based approach to SHM allows engineers to develop predictive models that can be used for O\&M decision-making, even in the absence of labelled data. This feat is accomplished by leveraging transfer-learning techniques to share information between structures within a population. Negative transfer is a widely acknowledged risk associated with transfer learning, whereby transfer degrades prediction performance for the target domain. In the context of PBSHM O\&M decision-making, negative transfer becomes a important phenomenon as it may result in an increase in unnecessary inspections, or structural failures -- both of which have costs associated. For this reason, it is important to be able to decide aspects such as when and what to transfer. To this end, the current paper has presented a general framework that facilitates the selection of optimal transfer strategies by considering the expected value of information transfer. In order to compute the expected value of information transfer, the framework leverages structural similarity to predict the post-transfer prediction quality; which is then mapped to a utility function defined with respect to O\&M decisions. While there are several limitations and considerations that must be recognised, the decision framework ultimately provides a means for reducing O\&M costs and improving structural safety by facilitating information sharing in a manner that avoids negative transfer.

%It was observed that for both cases, an abrupt amplitude change occurs when the temperature reaches a certain value. 

\vspace{24pt}
\noindent \uppercase{\textbf{Acknowledgment}} \vspace{12pt}

The authors would like to gratefully acknowledge the support of the UK Engineering and Physical Sciences Research Council (EPSRC) via grant references EP/W005816/1 and EP/R006768/1. For the purposes of open access, the authors have applied a Creative Commons Attribution (CC BY) license to any Author Accepted Manuscript version arising.

%------------------------------------------------------------------------------------------------------------
%                                                 BIBLIOGRAPHY
%------------------------------------------------------------------------------------------------------------
\vspace{24pt}

\small 

\bibliographystyle{iwshm}
\bibliography{IWSHM_AJH}

%------------------------------------------------------------------------------------------------------------
%------------------------------------------------------------------------------------------------------------
%------------------------------------------------------------------------------------------------------------

%\includepdf[pages={1}]{copyrightform.pdf}

\end{document}